\documentclass[journal]{IEEEtran}
\usepackage[utf8]{inputenc} % allow utf-8 input
\usepackage[T1]{fontenc}    % use 8-bit T1 fonts
\usepackage{hyperref}       % hyperlinks
\usepackage{url}            % simple URL typesetting
\usepackage{booktabs}       % professional-quality tables
\usepackage{amsfonts}       % blackboard math symbols
\usepackage{microtype}
\usepackage{subfigure}
\usepackage{epsfig}
\usepackage{graphicx}
\usepackage{amsmath}
\usepackage{amssymb}
\usepackage{subfigure}
\usepackage{float}

\ifCLASSINFOpdf
\else
\fi
\begin{document}
\title{Discovery Radiomics with CLEAR-DR: Interpretable Computer Aided Diagnosis\\ of Diabetic Retinopathy}

\author{Devinder Kumar,~\IEEEmembership{Student Member,~IEEE,}
        Graham W. Taylor,~\IEEEmembership{Member,~IEEE,}
        and Alexander Wong,~\IEEEmembership{Senior Member,~IEEE}% <-this % stops a space
\thanks{Devinder Kumar and Alexander Wong are with the University of Waterloo,
ON, N2L-3G1 Canada. E-mail: \{devinder.kumar, a28wong\}@uwaterloo.ca.}% <-this % stops a space
\thanks{Graham W. Taylor is with the University of Guelph, Canadian Institute of Advanced Research (CIFAR), and Vector Institute, Toronto, Canada. E-mail: gwtaylor@uoguelph.ca}% <-this % stops a space
}


\markboth{}%
{Kumar \MakeLowercase{\textit{et al.}}: CLEAR-DR: Interpretable Computer Aided Diagnosis of Diabetic Retinopathy}



\maketitle


\vspace{-2cm}
\begin{abstract}
Objective: Radiomics-driven Computer Aided Diagnosis (CAD) has shown considerable promise in recent years as a potential tool for improving clinical decision support in medical oncology, particularly those based around the concept of Discovery Radiomics, where radiomic sequencers are discovered through the analysis of medical imaging data. One of the main limitations with current CAD approaches is that it is very difficult to gain insight or rationale as to how decisions are made, thus limiting their utility to clinicians. Methods: In this study, we propose CLEAR-DR, a novel interpretable CAD system based on the notion of CLass-Enhanced Attentive Response Discovery Radiomics for the purpose of clinical decision support for diabetic retinopathy. Results: In addition to disease grading via the discovered deep radiomic sequencer, the CLEAR-DR system also produces a visual interpretation of the decision-making process to provide better insight and understanding into the decision-making process of the system. Conclusion: We demonstrate the effectiveness and utility of the proposed CLEAR-DR system of enhancing the interpretability of diagnostic grading results for the application of diabetic retinopathy grading. Significance: CLEAR-DR can act as a potential powerful tool to address the uninterpretability issue of current CAD systems, thus improving
their utility to clinicians.
\end{abstract}


% \begin{center}0 \bfseries EDICS Category: 3-BBND \end{center}
% \fi
\IEEEpeerreviewmaketitle

\section{Introduction}
\label{intro}

\begin{figure}[t]
\begin{center}
   \includegraphics[trim = 0cm 2.5cm 0cm 0cm ,height = 5cm,width=1\linewidth]{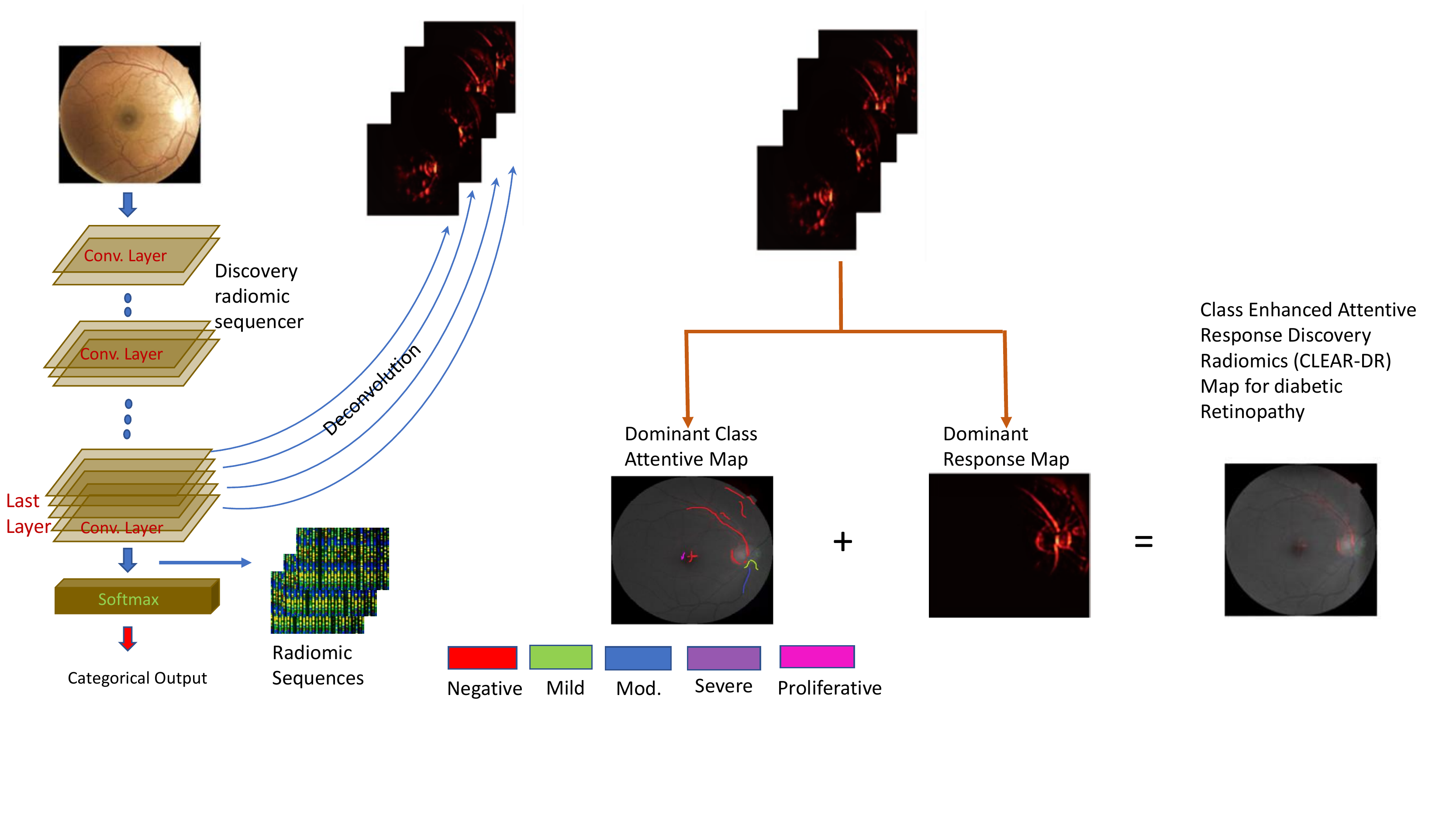}
\end{center}
\vspace{-0.5cm}
   \caption{An overview of the proposed \textbf{CL}ass-\textbf{E}nhanced \textbf{A}ttentive \textbf{R}esponse \textbf{D}iscovery \textbf{R}adiomics (CLEAR-DR) system for interpretable computer-aided diagnosis of diabetic retinopathy.  For a new patient case, the patient's retinal fundus image is fed into the discovered deep radiomic sequencer and individual attentive response maps are computed for each disease grade based on the last layer of the sequencer.  Based on this set of attentive response maps, two different maps are computed: 1) a dominant attentive response map, which shows the level of contribution of each location to the decision-making process, and 2) a dominant class attentive map, which shows the dominant grade associated with each location influencing the decision-making process. Finally, the dominant attentive response map and the dominant attentive class map are combined to produce the final CLEAR-DR map for a given retinal fundus image, thus enabling the clinician to visualize the factors taken by the proposed system in predicting disease grades.}
   \vspace{-0.5cm}
\label{fig:clap_explain}
\end{figure}

\begin{figure*}[t]
\begin{center}
   \includegraphics[trim = 0cm 6.5cm 0cm 0cm ,height = 5.cm,width=0.9\linewidth]{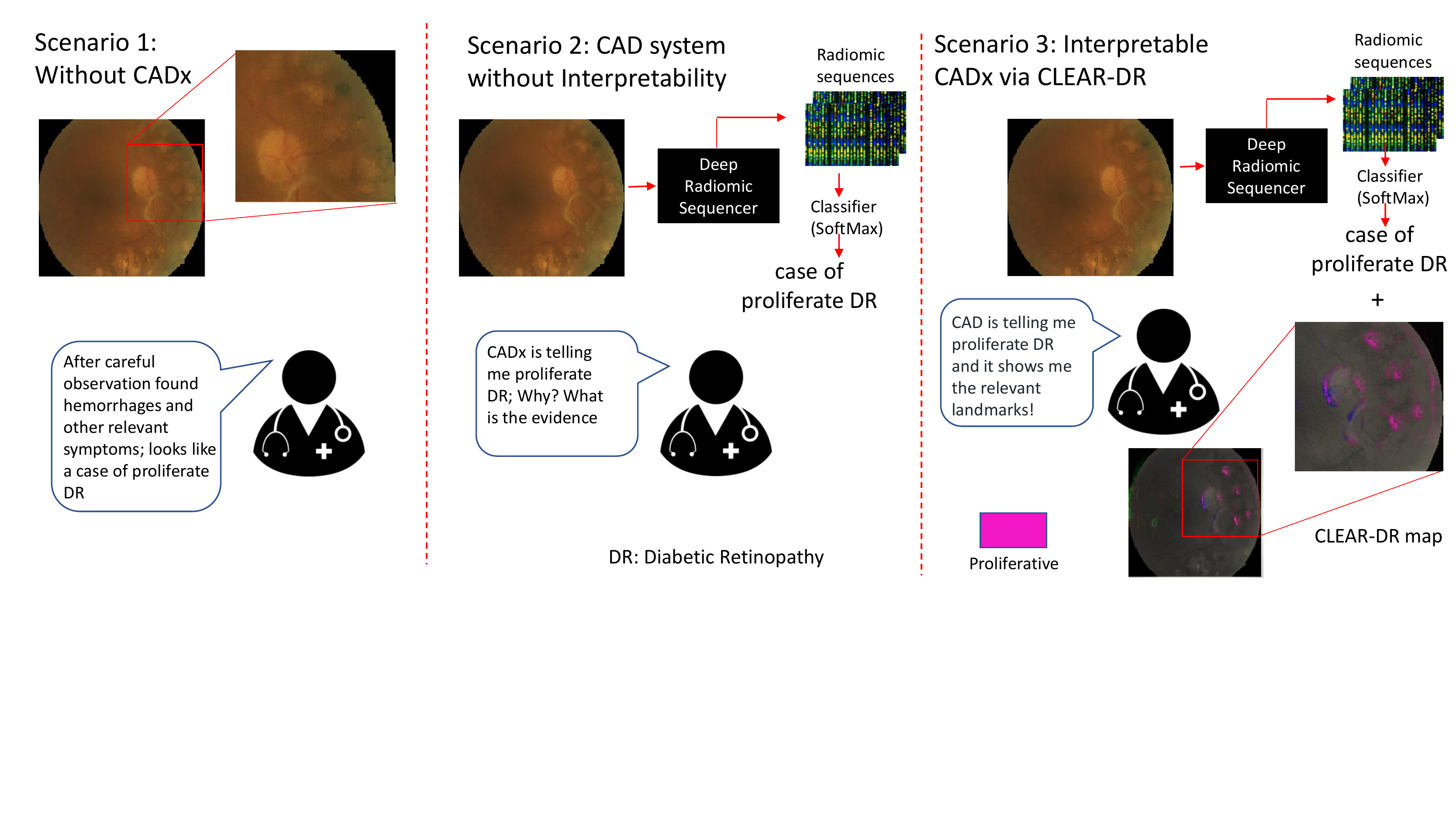}
\end{center}
   \caption{Three different scenarios for grading diabetic retinopathy: 1) without CAD, 2) CAD system without interpretability, 3) interpretable CAD via CLEAR-DR.  The proposed CLEAR-DR system improves clinical interpretability by providing effective visual interpretations of the decision-making process. The CLEAR-DR maps allows for the visualization of i) the attentive regions of interest responsible for grading decisions made by the deep radiomic sequencer; ii) their level of contribution to the grading decision; and iii) the dominant grade associated with each attentive region of interest. This visualization enables clinicians to better understand the rationale behind the grading decision made by the deep radiomic sequencer. }
      \vspace{-0.5cm}
\label{fig:motivation}
\end{figure*}
% by the rest of the first word in caps.0.3

%\vspace{-0.5cm}

Diabetic retinopathy is a medical condition that causes damage to the retina due to diabetes. It is the leading cause of blindness in the world. Traditionally, clinical diagnosis for diseases such as diabetic retinopathy is highly subjective based purely on a clinician's experience. As such, these diagnoses have high inter- and intra-observer variability. The prevalence of computer-aided diagnosis (CAD) systems to support clinicians in their decision-making process has risen, enabling faster and more accurate diagnostic decisions with lower variability. In particular, radiomics-driven CAD has become an increasingly more prevalent area of research focus, where radiomic sequences consisting of a large number of image-based features are extracted and used to help clinicians make more informed decisions, and provide a virtual second opinion~\cite{aerts2014decoding}. However, traditional radiomic sequences comprise largely of generic, hand-crafted features, which may be limiting in characterizing unique disease traits.

More recently, the concept of \textit{Discovery Radiomics} has been shown to be particularly promising for oncology decision support by directly discovering radiomic sequencers based on medical imaging data~\cite{kumar2015lung}, resulting in radiomic sequences that are tailored to characterizing unique disease traits. A particularly powerful use of Discovery Radiomics is for the discovery of deep radiomic sequencers, which leverage deep neural network (DNN) architectures to learn and extract subtle, latent features associated with key disease characteristics. Such DNN-based approaches have shown considerable promise in detecting diabetic retinopathy~\cite{gulshan2016development,takahashi2017applying}.

One of the main limitations of this approach is that it is very difficult to gain insight or rationale as to how decisions are made. This limits its utility to clinicians and hinders widespread adoption in clinical settings. Quantitative metrics such as accuracy or AUC do not convey any information about \emph{how} a particular deep radiomic sequencer makes decisions, thus they are often labeled as uninterpretable ``black boxes''. There is a vital need for complimentary decision support systems that can help and aid clinicians in understanding the decision making process of deep radiomic sequencers.

Some recent work pertaining to the interpretability and understanding of the inner working of DNNs has been developed in the context of convolutional neural networks for generic visual recognition tasks. For example, Zeiler and Fergus\cite{zeiler2014visualizing} proposed the use of an inverse parallel network known as a deconvolutional network paired with a CNN to visualize its features. The deconvolutional network re-projects the activations from the higher-level latent space back to the input such that they can be visualized in the image domain. Springenberg et al.~\cite{springenberg2014striving} proposed a visualization method known as guided backpropagation where only positive gradients were allowed to flow through backpropagation. In~\cite{zhou2016learning} Zhou et al.~proposed the class activation map (CAM) based on adding a global average pooling layer to the network to be visualized. CAMs are formed using only a single forward propagation though they only provide a rough estimation map of attentive regions. Montavon et. al.~\cite{montavon2017explaining} proposed a method based on Taylor expansion to improve the sharpness of the visualization. All of the these methods however, only highighlight the regions of attention and provide no meaning to the assignments other than that they should form a coherent set of interpretable pixels, primarily of convolutional, ReLU, and max-pooling layers. Recently, Gondal et.al.~\cite{gondal2017weakly} leveraged CAM maps to highlight the lesion areas for diabetic retinopathy; however, this approach provides no interpretation of grading information and thus is limited in providing clinical insight on grading decisions.
\newline Motivated by the need for clinical interpretability, we propose \textbf{CLEAR-DR}, a novel interpretable CAD system based on the notion of \textbf{C}Lass-\textbf{E}nhanced \textbf{A}ttentive \textbf{R}esponse \textbf{D}iscovery \textbf{R}adiomics for the purpose of clinical decision support for diabetic retinopathy. CLEAR-DR not only generates discriminative radiomic sequences for making grading decisions for diabetic retinopathy as a use case, but also visually interpret and understand these decisions via information back-propagation. The back-propagation is done through the discovered radiomic sequencer by embedding the CLEAR approach proposed by Kumar et. al.~\cite{kumar2017explaining}. This process is designed to enable grade-level interpretability. As shown in Fig.~\ref{fig:motivation}, CLEAR-DR can also help in reducing inter-observer variability and intra-observer variability while speeding up the overall diagnostic process. The main contribution of the proposed CLEAR-DR CAD system is as follows:
\begin{itemize}
\item To the best of the authors' knowledge, this is the first interpretable deep radiomic sequencer-driven CAD system proposed that enables the visualization of multi-class grading processes.
\item The study shows a direct qualitative correlation between the medically relevant landmarks that human experts use for diabetic retinopathy grading and the landmarks used by CLEAR-DR for classifying different grades of diabetics through retinopathy images.
\end{itemize}

\vspace{-0.3cm}
\section{Class Enhanced Attentive Response for Diabetic Retinopathy (CLEAR-DR)}
\label{approach}
\vspace{-0.1cm}
This section explains the procedure for creating the proposed interpretable CLEAR-DR CAD system. The procedure can be explained in two parts: First, discovery of a deep radiomic sequencer via Discovery Radiomics and second, creating the CLEAR-DR maps for visually interpreting the decision making process of the discovered deep radiomic sequencer. Both parts are explained below.
\vspace{-0.5cm}
\subsection{Discovering the Deep Radiomic Sequencer for Diabetic Retinopathy}
\begin{figure}[t!]
\begin{center}
   \includegraphics[trim = 1cm 10cm 2cm 4cm ,height = 3cm,width=1\linewidth]{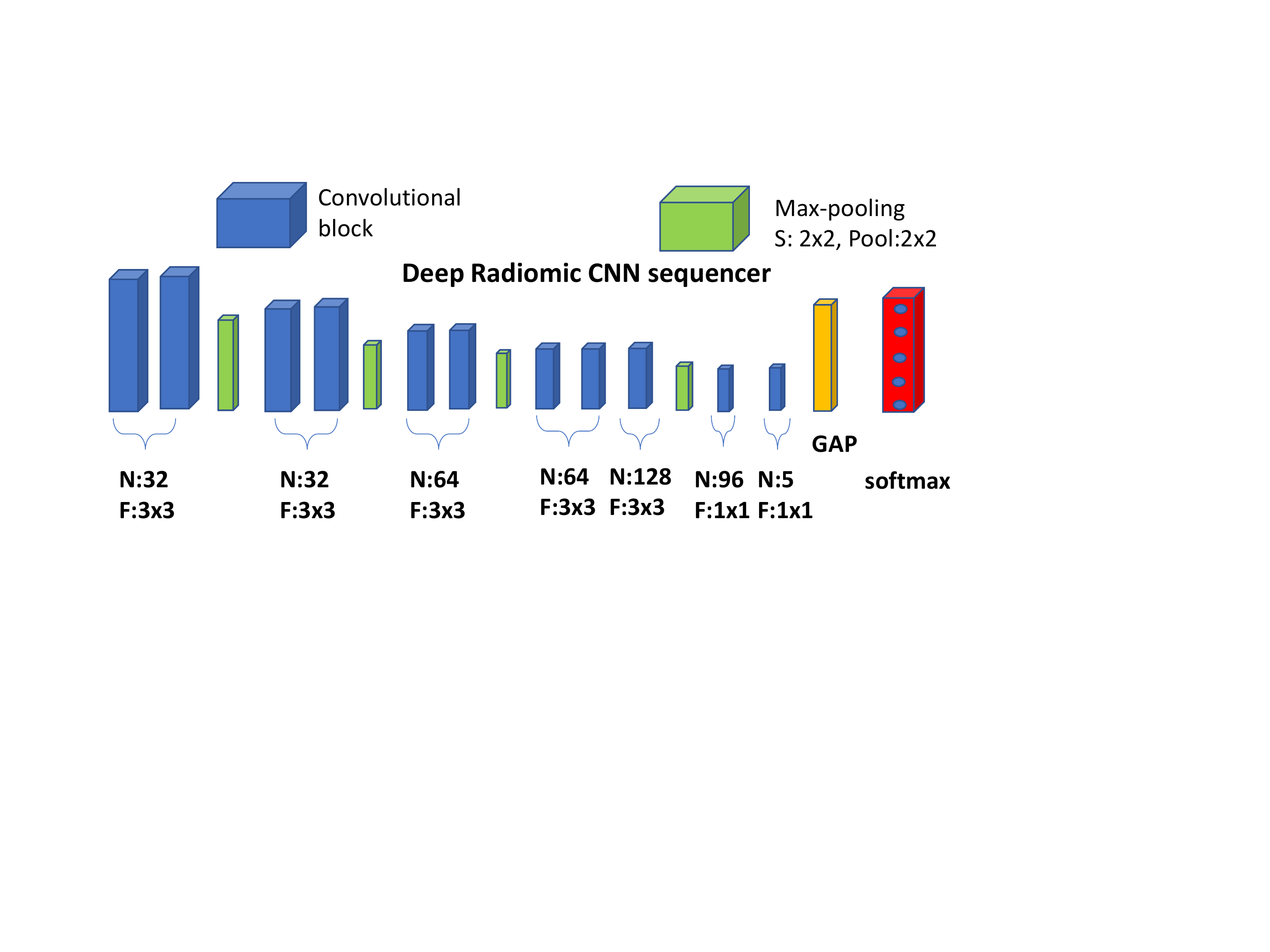}
\end{center}
   \caption{Architecture of the convolutional radiomic sequencer used in the deep radiomic sequencer discovery  process. The radiomic sequencer is embedded in the sequencer discovery process, which augments a set of fully convolutional layers, a rectified linear unit layer, max-pooling, global average pooling (GAP) and a loss layer at the end of the sequencer for the learning process.
}
 \vspace{-0.7cm}
\label{fig:sequencer}
\end{figure}

To discover the aforementioned deep convolutional radiomic sequencer, we construct a deep Convolutional Neural Network (CNN) architecture for the radiomic sequencer discovery process, where the radiomic sequencer is directly embedded in the sequencer discovery architecture and learned based on the available retinal fundus imaging data, and clinician-provided ground truth labels for diabetic retinopathy. An overview of the radiomic sequencer discovery architecture is shown in Fig.~\ref{fig:sequencer}. The receptive fields in the convolutional sequencing layers of the radiomic sequencer are learned in a supervised manner based on the input retinal fundus imaging data. This process allows us to learn specialized receptive fields in the custom radiomic sequencer that better characterize the unique diabetic retinopathy grading characteristics.
\vspace{-0.4cm}
\subsection{Creating the CLEAR-DR Map}
After the deep radiomic sequencer discovery, we create interpretable maps for CLEAR-DR system, called CLEAR-DR maps. CLEAR-DR maps present two sets of information. First, they present the regions in the diabetic retinopathy image, along with their level of influence, that are responsible for the decision made by the radiomic sequencer. Second, it presents the dominant grade associated with the above aforementioned regions. The procedure for creating the CLEAR-DR maps is shown in Fig.~\ref{fig:clap_explain}, which can be explained as follows: First, individual attentive response maps are computed for each kernel associated with a grade by back-projecting activations from the output layer of the deep radiomic sequencer. Based on this set of attentive response maps, two different types of maps are computed: 1) a dominant attentive response map, which shows the dominant attentive level for each location in the image; and 2) a dominant class attentive map, which shows the dominant grade involved in the decision-making process at each location. Finally, the dominant attentive response map and the dominant attentive grade map are combined visually by using color and intensity to produce the final CLEAR-DR map.

As the proposed method extends upon the CLEAR approach, for better understanding we keep the same notation as described by Kumar et.al.~\cite{kumar2017explaining}.
The first step of CLEAR-DR map is to compute a set of individual attentive response maps, which we will denote as $\left\{R(\underline{x}\lvert d)|1 \leq d \leq N\right\}$, where $N$ is the number of grades in diabetic retinopathy. This is achieved by back-propagating the responses of each kernel in the last convolutional layer of the deep radiomic sequencer. To explain the formulation, first consider a single layer of a deep radiomic sequencer. Let $\hat h_l$ be the deconvolved output response of the single layer $l$ with $K$ kernel weights $w$. The deconvolution output response at layer $l$ then can be then obtained by convolving each of the feature maps $z_{l}$ with kernels $w_{l}$ and summing them as: $\hat h_{l} = \sum_{k=1}^K z_{k,l} * w_{k,l} .$ Here $*$ represents the convolution operation. For notational brevity, we can combine the convolution and summation operation for layer $l$ into a single convolution matrix $G_{l}$. Hence the above equation can be denoted as: $\hat h_{l} = G_{l}z_{l}$.

For multi-layered sequencers, we can add an additional un-pooling operation $U$. Thus, we can calculate the deconvolved output response from feature space to input space for any layer $l$ in a multi-layer  sequencer as:
\vspace{-0.2cm}
\begin{equation}
 R_{l} = G_{1}U_{1}G_{2}U_{2} ....G_{l-1}U_{l-1}G_{l}z_{l}
\end{equation}
\vspace{-0.5cm}

For CLEAR-DR maps, we specifically calculate the output responses from individual kernels of the last layer of the sequencer. Hence, given a network with last layer $L$ containing $K=N$ kernels, we can calculate the attentive response map; $R(\underline{x}\lvert c)$ (where $\underline{x}$ denotes the response back-projected to the input layer, and thus an array the same size as the input) for any grade-specific kernel $c$ (${1 \leq c \leq N}$) in the last layer as:
\vspace{-0.1cm}
\begin{equation}
 {R(\underline{x}\lvert d)} = G_{1}U_{1}G_{2}U_{2} ....G_{L-1}U_{L-1}G_{L}^d z_{L} .
\end{equation}

Here $G_{L}^d$ represents the convolution matrix operation in which the kernel weights $w_{L}$ are all zero except that at the $d$\textsuperscript{th} location.

Given the set of individual attentive response maps, we then compute the dominant attentive class map, $\hat{C}(\underline{x})$, by finding the class at each pixel that maximizes the attentive response level, $R(\underline{x}\lvert d)$, across all classes:
\vspace{-0.1cm}
\begin{equation}
\vspace{-0.2cm}
 \hat{C}(\underline{x}) = \operatornamewithlimits{argmax}\limits_{c} {R(\underline{x} \lvert d)} .
\end{equation}
\vspace{-0.2cm}

Given the dominant attentive class map, $\hat{C}(\underline{x})$, we can now compute the dominant attentive response map, $D_{\hat{C}}(\underline{x})$, by selecting the attentive response level at each pixel based on the identified dominant grade, which can be expressed as follows:
\vspace{-0.1cm}
\begin{equation}
 D_{\hat{C}}(\underline{x}) = R(\underline{x}\lvert \hat{C}) .
\end{equation}

To form the final CLEAR-DR map, we map the dominant grade attentive map and the dominant attentive response map in the HSV color space as follows:
\vspace{-0.1cm}
\begin{equation}
   H  = F(\hat{C}(\underline{x})) ;
   S  = 1 ;
   V  = D_{\hat{C}}(\underline{x})
\end{equation}

Here $F(.)$ is the color map dictionary that assigns an individual color to each dominant attentive grade, $d$. Fig.~\ref{fig:clap_explain} shows an example of the CLEAR map overlayed on the image.

%-----------------------------------------------------------------------

%-----------------------------------------------------------------------

\section{Experiments and Results}
\label{exp}
This section presents the experimental setup and the qualitative experiments performed to show the efficacy of the CLEAR-DR maps via the discovery radiomics framework. We conducted experiments on the Kaggle diabetic retinopathy dataset~\cite{dataset2015kaggle} using a CNN-based deep radiomic sequencer. Details about the dataset, training and creation of CLEAR-DR maps are explained below.
\vspace{-0.5cm}

\begin{figure}[t]
\begin{center}
   \includegraphics[trim = 0cm 7cm 0cm 1cm ,height = 3cm,width=1\linewidth]{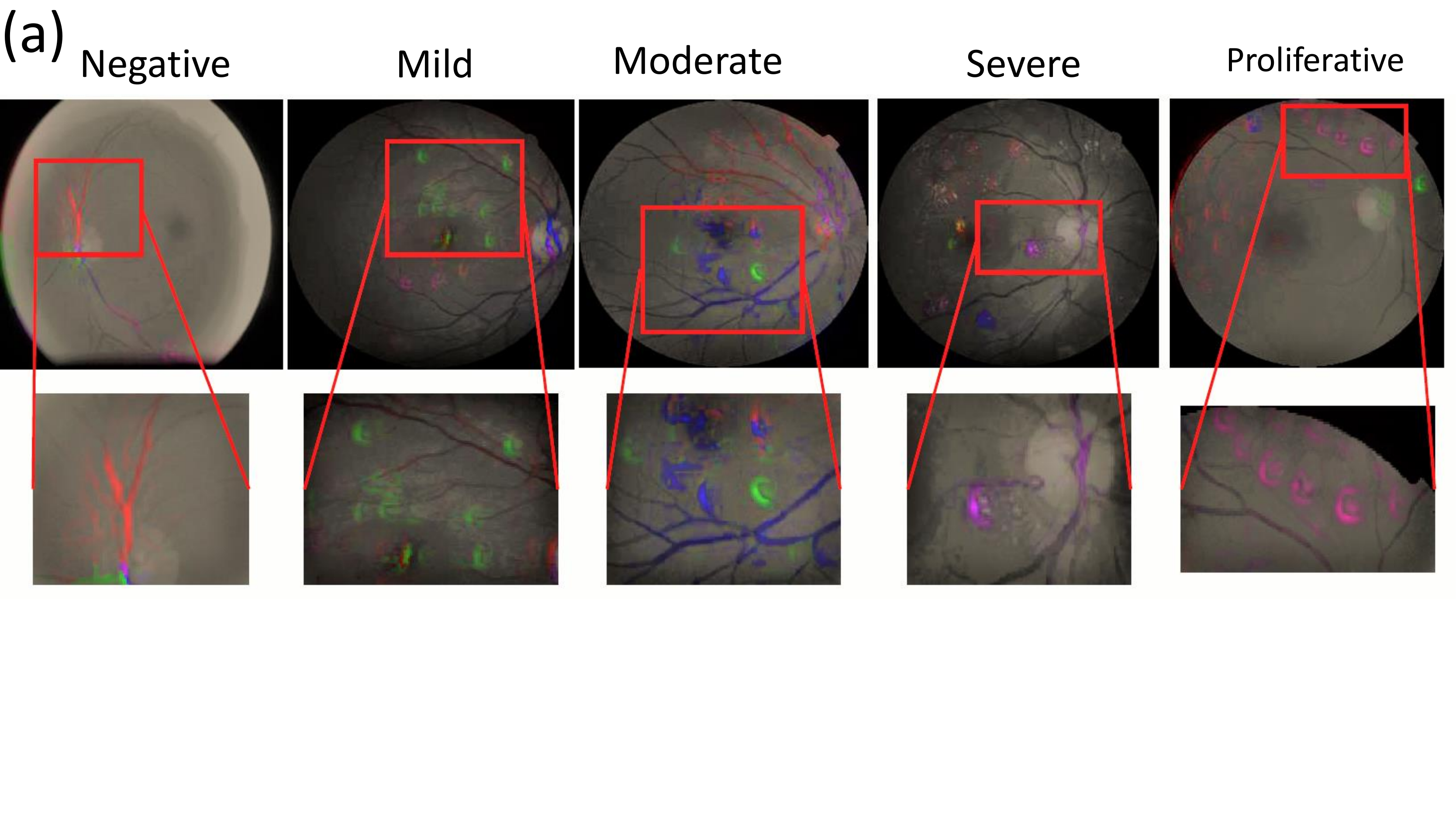}
\end{center}
\begin{center}
   \includegraphics[trim = 0cm 03cm 0cm 1cm ,height =4cm,width=1\linewidth]{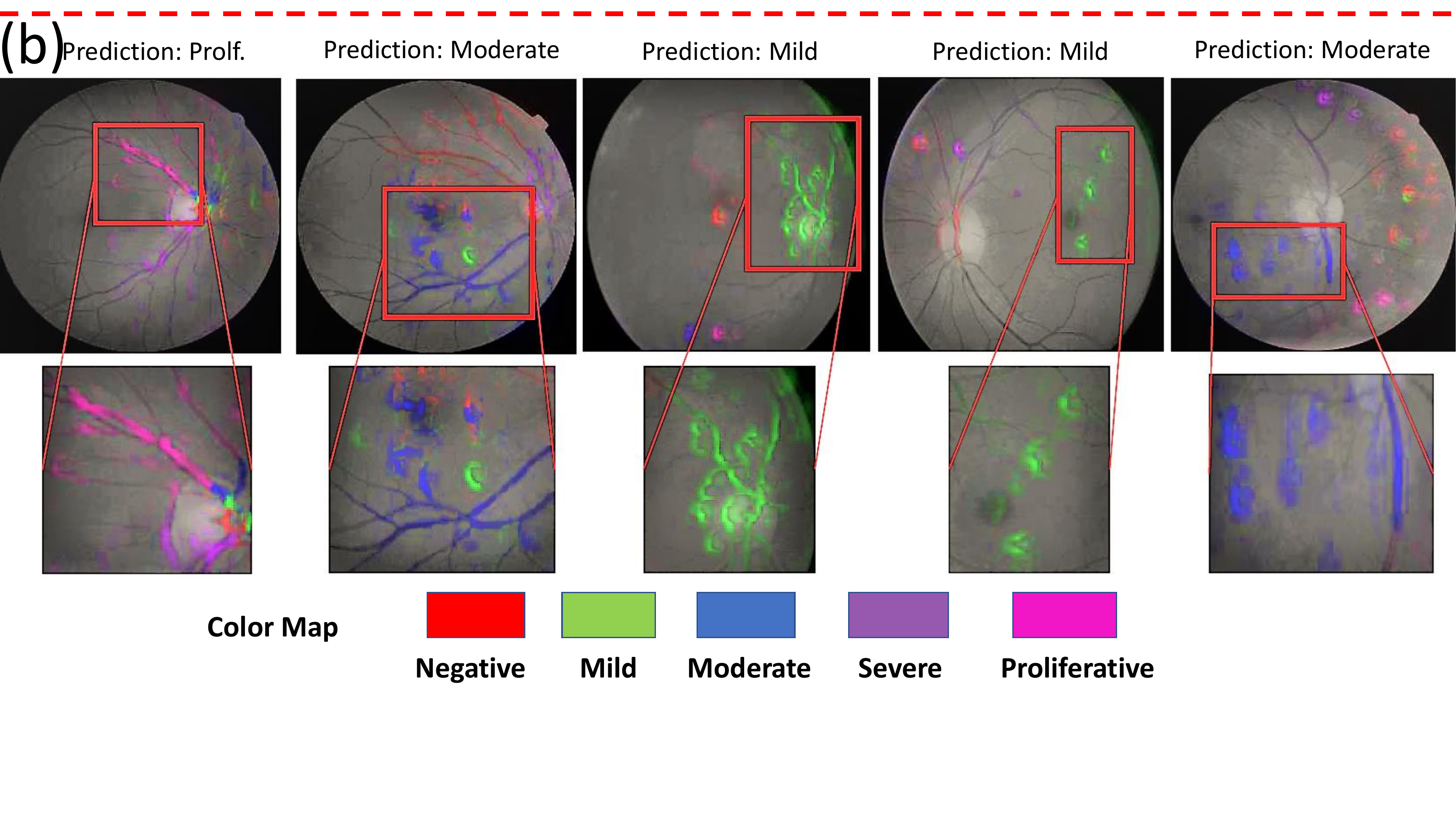}
\end{center}
\vspace{-0.4cm}
   \caption{Correctly (a) and Mis-classified (b) examples for all diabetic retinopathy grades. Each color represents a single grade, as identified  by the color map at the bottom of the figure. As well, the red box indicates the most attentive region used for grade prediction. It can be observed that the attentive regions used by the deep radiomic sequencer for making correct decisions corresponds to medically relevant landmarks, thus providing additional evidence for the proposed prediction. Best viewed in color and zoomed in.}
\vspace{-0.4cm}
\label{fig:results}
\end{figure}

\subsection{Kaggle Diabetic Retinopathy Dataset}
The Kaggle diabetic retinopathy dataset~\cite{dataset2015kaggle} consists of high-resolution retinal fundus images with varying degrees of illumination conditions captured using different types of cameras. The retinal fundus images in the dataset were clinically annotated with five different grades related to the presence of diabetic retinopathy. The five grades of diabetic retinopathy are as follows: 0: Negative, 1: Mild, 2: Moderate, 3: Severe, and 4: Proliferative. The dataset consists of images from both right and left eyes. Mild noise is present in both the images and ground truth labels.
\vspace{-0.3cm}
\subsection{Experimental Setup: Training Discovery Radiomics Sequencer}
To create and train a deep radiomic sequencer for diabetic retinopathy, we use a CNN as shown in Fig.~\ref{fig:sequencer}. To train this deep radiomic sequencer, we selected retinal fundus images for one eye (right) only and performed an automatic selective cropping to remove the background information. The use of a single eye led to 53,354 images in total. For evaluation purposes, we divided the dataset into 90\% and 10\% of the dataset for training and testing respectively. We also performed horizontal and vertical flipping along with channel-wise normalization for the whole dataset as data augmentation. Using the above setup, we trained the deep radiomic sequencer and achieved an accuracy of 73.2\% overall. It is important to note here that the goal of this study is to create an interpretable system for diabetic retinopathy, and thus the focus is on interpretability of grading decisions made using the deep radiomic sequencer.  As such, the accuracy of the proposed CAD system can be improved further by leveraging alternative DNN architectures and other optimization approaches.
\vspace{-0.3cm}
\subsection{Qualitative Experiments: CLEAR-DR for Interpretable CAD}
To demonstrate the efficacy of interpretability with the CLEAR-DR system, we took the above discovered deep radiomic sequencer and created CLEAR-DR maps using the procedure shown in Fig.~\ref{fig:clap_explain} for all diabetic retinopathy grades for both scenarios i.e., cases where the CLEAR-DR CAD system correctly predicts a diabetic's grade (Fig.~\ref{fig:results}(a)) and cases where it failed to identify the correct grade (Fig.~\ref{fig:results}(b)). Observing the individual cases in Fig.~\ref{fig:results}, it is evident that CLEAR-DR maps are able to explain both scenarios i.e., where the grade was either correctly or mis-classified. For both scenarios, it provides the attention areas thus giving a rationale in both cases for particular decisions that the clinician can reason with.

Other specific observations can also be made from Fig.~\ref{fig:results}. For example, in the correctly classified diabetic retinopathy case in  Fig.~\ref{fig:results}(a), it can be observed that the deep radiomic sequencer mainly focuses on the veins near the eye-balls in the retinal fundus image as there is an absence of abnormalities. A similar observation about the abnormality and the extent of it can be made in other correctly identified cases. In the mis-classified case of mild diabetic retinopathy in Fig.~\ref{fig:results} (b), it can be seen the deep radiomic sequencer fails to focus on the correct abnormalities in the though it identifies the abnormality, it considers it as proliferative case.

Based on these results and observations, it is evident that CLEAR-DR maps show a direct correlation between the medically relevant landmarks for identifying the condition of diabetic retinopathy and the attentive areas used by the CLEAR-DR CAD system for grading diabetic retinopathy. Thus, we argue that the CLEAR-DR maps are effective for understanding and interpreting the classification decisions made by a CAD system and also for providing a reason for their decision making process in clinical settings.

\section{Conclusion}
\label{conclusion}
In this study, we proposed CLEAR-DR, an interpretable CAD system via Discovery Radiomics. Using the Class-Enhanced Attentive Response (CLEAR) approach, we created a system based on discovery of a deep radiomic sequencer that not only generates discriminating radiomic sequences for diabetic retinopathy grading but also provides a mechanism to visually interpret its decision making process. Qualitative experiments show a direct correlation between the medically relevant landmarks used by human experts for grading and the visual features identified and used by the CLEAR-DR system for diabetic retinopathy grading. Thus, the proposed approach has great potential to reduce inter- and intra-observer variability and to accelerate the overall screening and diagnosis process while improving consistency and accuracy in clinical settings.

{\small
\bibliographystyle{ieee}
\bibliography{egbib}
}

\end{document}